\documentclass[11pt]{article}
\usepackage[utf8]{inputenc}
\usepackage[T1]{fontenc}
\usepackage{lmodern}
\usepackage[margin=1in]{geometry}
\usepackage{microtype}
\usepackage{amsmath}
\usepackage{amssymb}
\usepackage{amsthm}
\usepackage{bbm}
\usepackage{graphicx}
\usepackage{xcolor}
\usepackage{algorithm}
\usepackage{algpseudocode}
\usepackage[colorlinks=true,linkcolor=blue!60!black,citecolor=blue!60!black,urlcolor=blue!60!black]{hyperref}

\newtheorem{theorem}{Theorem}
\newtheorem{lemma}{Lemma}
\theoremstyle{remark}
\newtheorem{remark}{Remark}

\newcommand{\E}{\mathbb{E}}
\newcommand{\indic}{\mathbbm{1}}
\newcommand{\bp}{\mathbf{p}}

\title{Greedy Resource Allocation for Online Estimation\\ of Independent Bernoulli Processes}
\author{Yash Satsangi}
\date{}

\begin{document}

\maketitle

\begin{abstract}
We consider the problem in which an agent must select informative observations under resource constraints to continuously monitor a set of hidden variables in an online setting. Existing approaches model this problem as a partially observable Markov decision process and use planning methods to solve it. However, these approaches assume either a model of the world or a simulator, and they do not characterize the optimal policy for learning online. To this end, we abstract the problem to the simple case of estimating the parameters of $n$ independent Bernoulli distributions when, at each time step, the agent observes only the outcome it chooses. We show that minimizing the cumulative mean squared error is equivalent to maximizing a cumulative reward that we derive, and our main result shows that long-term reasoning is not required: the policy that is greedy with respect to this reward is optimal. Since the reward depends on the unknown parameters, we build on this result to develop an upper-confidence-bound (UCB) policy and bound, at each round, the gap between the reward of its action and that of the greedy action. Experiments show that the UCB policy and $\epsilon$-greedy both outperform random and round-robin selection.
\end{abstract}

\section{Introduction}
This paper studies the problem of learning the biases of $n$ independent coins when, at each time step, the learning agent can observe the outcome of only one coin of its choosing. The agent's aim is to estimate the biases of all coins at every time step, and at each time step it suffers an expected loss that is a function of its estimates and the true biases.

The motivation for this problem comes from information gathering and monitoring tasks, such as sensor selection and cooperative active perception \cite{spaan2008cooperative, spaan2015decision, bajcsy2018revisiting, satsangi2018exploiting}, in which one or more agents gather information so that another agent, typically a human, can monitor the state of the world and make informed decisions. For example, consider a set of agents monitoring a shopping mall to detect suspicious activity to which a human must respond \cite{spaan2015decision, satsangi2018exploiting}. The agents only collect information; the decision of how to respond lies with the human.

The most common approach to such tasks is decision-theoretic planning \cite{spaan2015decision, satsangi2018exploiting}, in which a policy for action selection is computed from a model of the world that is either assumed or learned from historical data. Such tasks are typically modeled as a partially observable Markov decision process (POMDP) \cite{kaelbling1998planning, spaan2015decision} or a decentralized POMDP \cite{oliehoek2016concise}, and a planning method computes the policy that the agent then follows. While this approach is effective when a good model of the environment is available, it does not let the agent adapt to, or learn from, the data it collects while executing the task, and it suffers when the model is inaccurate. It also cannot fully exploit recent advances in learning algorithms, such as deep reinforcement learning \cite{mnih2013qlearning} and online learning \cite{slivkins2019introduction, seznec2019rotting, azar2017minimax}. Approaches that apply learning algorithms, including deep reinforcement learning, to such problems \cite{linke2020adapting, satsangi2020maximizing} focus on the application and do not establish whether these algorithms are optimal in principle for monitoring a hidden state.

This paper focuses on learning a policy for such tasks online, while the agent executes the task, with a theoretical grounding for the solution. We restrict the state of the world to $n$ independent Bernoulli distributions with static hidden parameters $p_a$, $a \in \{1, 2, \dots, n\}$. At each time step, the world tosses the $n$ coins, the agent estimates their biases, and the agent chooses which coin's outcome to observe. The agent estimates the biases from the history of its observations and, at each time step, suffers a loss that depends on its estimates and the true biases.

Motivated by frameworks from online learning and reinforcement learning such as multi-armed bandits \cite{auer2002finite, slivkins2019introduction} and rotting bandits \cite{levine2017rotting, seznec2019rotting}, we first convert loss minimization into reward maximization. Applying existing algorithms is not straightforward, however, because the agent's choice at each step affects its future rewards, so in general the agent must reason about the long-term consequences of its choices. Our contributions are as follows.
\begin{itemize}
    \item We show that minimizing the cumulative expected mean squared error is equivalent to maximizing a cumulative reward that we derive in closed form (Theorem~\ref{th:equivalence}).
    \item We prove that the policy that is greedy with respect to this reward is optimal, so no long-term reasoning is needed (Theorem~\ref{th:greedy}), and relate this result to classical Neyman and marginal allocation.
    \item Building on this result, we propose a UCB policy (Algorithm~\ref{alg:ucb}) and bound, at each round, the gap between the reward of its action and that of the greedy action (Theorem~\ref{th:ucb}).
    \item We compare the UCB policy, $\epsilon$-greedy, round-robin, and random selection empirically.
\end{itemize}

\section{Related Work}
Problem formulations related to ours have appeared in \cite{spaan2008cooperative, spaan2012partially, spaan2015decision, satsangi2018exploiting}. These works consider a more complex, practical setting in which an agent selects a set of sensors from which to receive observations in order to monitor the hidden state of the world, typically modeled as a POMDP \cite{kaelbling1998planning}. The methods they propose are planning methods that assume an accurate model of the world is available. In contrast, we focus on learning the selection policy, and we study a simpler setting in order to characterize the optimal learning policy in principle. Methods that address learning in such POMDPs with deep reinforcement learning \cite{linke2020adapting, satsangi2020maximizing} likewise do not focus on the optimality of the resulting algorithms.

Our setting is closely related to reward maximization in multi-armed bandits \cite{slivkins2019introduction} and to pure exploration in multi-armed bandits \cite{bubeck2010}. The main difference is that our agent performs two actions at every step: it selects an observation to learn from, and it predicts the biases of all $n$ coins. In the bandit setting, by contrast, the agent either trades off exploration and exploitation to collect reward from the best arm or, in pure exploration, answers queries about the best arm after exploring.

Another closely related setting is estimating the parameters of random vectors from bandit feedback \cite{ghosh2025}. There, too, the aim differs: the agent must answer queries at the end of its interaction with the arms, as in pure exploration, rather than continuously throughout. The results most relevant to ours are those on decaying and rotting bandits \cite{heidari2016tight, levine2017rotting, seznec2019rotting}. Our reward formulation was motivated by the possibility of using these results, but, as we discuss in Section~\ref{sec:greedypolicy}, they do not apply directly to our setting.

Closest to our setting is active learning for estimating the means of several arms \cite{antos2010active, carpentier2011ucb} and adaptive stratified sampling for Monte Carlo integration \cite{carpentier2011finite}. There, an agent allocates a fixed budget of $n$ samples among arms with unknown variances to estimate their means accurately, either uniformly well \cite{antos2010active, carpentier2011ucb} or through a weighted average \cite{carpentier2011finite}. \cite{aznag2023active} study the same allocation problem for the $p$-norm of the vector of estimator variances; for $p = 1$ this is the sum of mean squared errors that we consider at each time step. Their Variance-UCB algorithm attains regret $\tilde{O}(T^{-2})$ with a matching lower bound. These works measure the error at the final time $n$ (or $T$), and their algorithms take the budget as an input to set their confidence levels; \cite{carpentier2011finite} leave an anytime version, which does not require knowledge of $n$, to future work. We instead measure the error at every time step and seek a policy that is good throughout, which is the natural objective when the agent continuously reports its estimates to another agent. When the variances are known, the optimal static allocation for the sum of mean squared errors is classical Neyman allocation \cite{neyman1934two}, proportional to the standard deviations, and its exact integer version can be computed by a greedy, marginal-allocation procedure \cite{fox1966discrete, wright2012equivalence}; we discuss this connection after Theorem~\ref{th:greedy}.

Finally, \cite{medin2018optimal} studies a similar aim in a Bayesian setting, with optimal stopping rules as the main result, and does not address regret in an online setting. Our setting is inspired by monitoring tasks in which an agent continuously assists another agent in tracking a hidden state; to the best of our knowledge, this paper is the first to study such a setting from the perspective of online learning.

\section{Background}
We briefly review maximum-likelihood estimation for the Bernoulli distribution \cite{BishopBook}. These facts are used in the proof of Theorem~\ref{th:equivalence} and in Section~\ref{sec:ucb}.

A Bernoulli random variable $Z \in \{0,1\}$ with bias $p \in [0,1]$ models the outcome of flipping a coin: $Z = 1$ with probability $p$ and $Z = 0$ with probability $1-p$. Its probability mass function is
\begin{equation*}
    \Pr(Z = z) = p^z(1-p)^{1-z}, \quad z \in \{0,1\}.
\end{equation*}

\paragraph{Estimate.} Let $Z_1, Z_2, \dots, Z_s$ be independent samples of $Z$, where the bias $p$ is unknown. The log-likelihood of the samples is
\begin{equation*}
    \log \mathcal{L}(p) = \sum_{i=1}^{s} \big[ Z_i \log p + (1 - Z_i) \log (1-p) \big].
\end{equation*}
When $0 < \sum_{i=1}^{s} Z_i < s$, this function is concave in $p$, and setting its derivative to zero,
\begin{equation*}
    \frac{d}{dp} \log \mathcal{L}(p) = \frac{\sum_{i=1}^{s} Z_i}{p} - \frac{s - \sum_{i=1}^{s} Z_i}{1-p} = 0,
\end{equation*}
gives the maximum-likelihood estimate, which is the sample mean:
\begin{equation*}
    \hat{p} = \frac{1}{s} \sum_{i=1}^{s} Z_i.
\end{equation*}
When $\sum_{i=1}^{s} Z_i \in \{0, s\}$, the likelihood is maximized at $p = 0$ or $p = 1$, which the same formula gives.

\paragraph{Mean and variance.} Each sample has $\E[Z_i] = p$ and $\mathrm{Var}(Z_i) = p(1-p)$. By linearity of expectation and independence of the samples,
\begin{equation*}
    \E[\hat{p}] = \frac{1}{s} \sum_{i=1}^{s} \E[Z_i] = p, \qquad
    \mathrm{Var}(\hat{p}) = \frac{1}{s^2} \sum_{i=1}^{s} \mathrm{Var}(Z_i) = \frac{p(1-p)}{s}.
\end{equation*}
That is, $\hat{p}$ is unbiased.

\paragraph{Mean squared error.} The mean squared error of an estimator is the sum of its squared bias and its variance. Since $\hat{p}$ is unbiased,
\begin{equation} \label{eq:mse}
    \E\big[(p - \hat{p})^2\big] = \big(\E[\hat{p}] - p\big)^2 + \mathrm{Var}(\hat{p}) = \frac{p(1-p)}{s}.
\end{equation}
Finally, $p(1-p) \leq \frac{1}{4}$ for all $p \in [0,1]$, with equality at $p = \frac{1}{2}$, and $p(1-p) = 0$ for $p \in \{0, 1\}$. Coins with bias near $\frac{1}{2}$ are therefore the hardest to estimate.

\section{Problem Setting} \label{sec:setting}
Let $[k]$ denote the set $\{1, 2, \dots, k\}$ for any positive integer $k$. There are $n$ coins (also called \emph{arms} or \emph{actions}) with unknown biases $\bp = \langle p_1, p_2, \dots, p_n \rangle$, and we write $\sigma^2_a = p_a(1-p_a)$ for the variance of coin $a$. At each time step $t \in [h]$, where $h$ is the horizon:
\begin{enumerate}
    \item the environment tosses all $n$ coins; the outcome of coin $a$ is $z_{a,t} \in \{0,1\}$, drawn independently with $\Pr(z_{a,t} = 1) = p_a$;
    \item the agent predicts the biases $\langle \hat{p}_{1,t-1}, \dots, \hat{p}_{n,t-1} \rangle$ from its observations so far and suffers the loss $\sum_{a \in [n]} (p_a - \hat{p}_{a,t-1})^2$;\footnote{We use the squared error; other losses, such as the entropy of the hidden variable, could also be considered (Section~\ref{sec:extensions}).}
    \item the agent selects an arm $a_t \in [n]$ and observes $z_{a_t,t}$.
\end{enumerate}
A policy $\pi$ maps the history of actions and observations to the next action, $a_t = \pi(\mathcal{H}_{t-1})$, where
\begin{equation*}
    \mathcal{H}_{t-1} = \langle a_1, z_{a_1,1}, a_2, z_{a_2,2}, \dots, a_{t-1}, z_{a_{t-1},t-1} \rangle.
\end{equation*}
Let $N_{a,t}$ be the number of observations of arm $a$ after $t$ time steps. We assume that each arm is observed once before the first time step, so that $N_{a,0} = 1$ and
\begin{equation*}
    N_{a,t} = 1 + \sum_{t' \in [t]} \indic(a_{t'} = a).
\end{equation*}
The agent's prediction $\hat{p}_{a,t}$ is the maximum-likelihood estimate of $p_a$, that is, the mean of the $N_{a,t}$ observations of arm $a$ (including the initial one) available after $t$ time steps. Over the horizon, the agent suffers the expected loss
\begin{equation*}
    L(\bp, h, \pi) = \E\left[ \sum_{t \in [h]} \sum_{a \in [n]} (p_a - \hat{p}_{a,t-1})^2 \right].
\end{equation*}
The aim is to find a policy that minimizes this loss,
\begin{equation*}
    \pi^* \in \arg\min_{\pi \in \Pi_h} L(\bp, h, \pi),
\end{equation*}
where $\Pi_h$ denotes the set of policies for horizon $h$.

\section{From Loss Minimization to Reward Maximization}
The loss of the agent decreases as it collects observations, which suggests using results from decaying bandits \cite{heidari2016tight} and rotting bandits \cite{seznec2019rotting}. To do so, we first convert loss minimization into reward maximization.

In this section and the next, we analyze policies as fixed action sequences $\pi = \langle a_1, a_2, \dots, a_h \rangle$, so that the counts $N_{a,t}$ are deterministic. This is the benchmark relevant to an agent that knows $\bp$: the optimal sequence depends only on $\bp$, and we show in Section~\ref{sec:greedypolicy} that it is the greedy sequence (see Remark~\ref{rem:adaptive}).

For such a policy, let $f_a^{\pi}(\ell)$ denote the time step at which arm $a$ is observed for the $\ell$-th time, where $f_a^{\pi}(1) = 0$ is the initial observation, and set $f_a^{\pi}(N_{a,h}+1) = h$ for convenience.

\begin{theorem} \label{th:equivalence}
For each time step $t \in [h]$ and arm $a \in [n]$, define the reward of selecting $a$ at time $t$ as
\begin{equation*}
    R_t(a) \triangleq \frac{\sigma^2_a\, (h - t)}{(N_{a,t-1} + 1)\, N_{a,t-1}},
\end{equation*}
and the value of a policy $\pi = \langle a_1, \dots, a_h \rangle$ as
\begin{equation*}
    V^{\pi}_h = \sum_{t \in [h]} R_t(a_t).
\end{equation*}
Then
\begin{equation} \label{eq:loss-value}
    L(\bp, h, \pi) = h \sum_{a \in [n]} \sigma^2_a - V^{\pi}_h,
\end{equation}
so a policy maximizes $V^{\pi}_h$ if and only if it minimizes $L(\bp, h, \pi)$. Moreover,
\begin{equation} \label{eq:value}
    V^{\pi}_h = \sum_{a \in [n]} \sum_{\ell=2}^{N_{a,h}} \frac{\sigma^2_a \big(h - f_a^{\pi}(\ell)\big)}{\ell(\ell-1)}.
\end{equation}
\end{theorem}
\begin{proof}
\emph{Step 1: loss at each time step.} The prediction at time $t$ is the mean of the $N_{a,t-1}$ observations of arm $a$ in $\mathcal{H}_{t-1}$, which are independent samples of coin $a$. By Equation~\eqref{eq:mse},
\begin{equation*}
    \E\big[(p_a - \hat{p}_{a,t-1})^2\big] = \frac{\sigma^2_a}{N_{a,t-1}}, \quad \text{so} \quad L(\bp, h, \pi) = \sum_{a \in [n]} \sum_{t \in [h]} \frac{\sigma^2_a}{N_{a,t-1}}.
\end{equation*}

\emph{Step 2: group the loss by observation count.} Arm $a$ has exactly $\ell$ observations at the prediction steps $t \in \{f_a^{\pi}(\ell)+1, \dots, f_a^{\pi}(\ell+1)\}$, that is, for $f_a^{\pi}(\ell+1) - f_a^{\pi}(\ell)$ steps. Hence
\begin{equation*}
    \sum_{t \in [h]} \frac{\sigma^2_a}{N_{a,t-1}} = \sigma^2_a \sum_{\ell=1}^{N_{a,h}} \frac{f_a^{\pi}(\ell+1) - f_a^{\pi}(\ell)}{\ell}.
\end{equation*}

\emph{Step 3: telescope.} For every $\ell \geq 1$,
\begin{equation*}
    \frac{1}{\ell} = 1 - \sum_{u=2}^{\ell} \frac{1}{u(u-1)},
\end{equation*}
since $\frac{1}{u(u-1)} = \frac{1}{u-1} - \frac{1}{u}$. Substituting this and exchanging the order of summation,
\begin{align*}
    \sum_{\ell=1}^{N_{a,h}} \frac{f_a^{\pi}(\ell+1) - f_a^{\pi}(\ell)}{\ell}
    &= \sum_{\ell=1}^{N_{a,h}} \big(f_a^{\pi}(\ell+1) - f_a^{\pi}(\ell)\big)
     - \sum_{u=2}^{N_{a,h}} \frac{1}{u(u-1)} \sum_{\ell=u}^{N_{a,h}} \big(f_a^{\pi}(\ell+1) - f_a^{\pi}(\ell)\big) \\
    &= h - \sum_{u=2}^{N_{a,h}} \frac{h - f_a^{\pi}(u)}{u(u-1)},
\end{align*}
where both inner sums telescope, using $f_a^{\pi}(1) = 0$ and $f_a^{\pi}(N_{a,h}+1) = h$. Summing over arms,
\begin{equation*}
    L(\bp, h, \pi) = h \sum_{a \in [n]} \sigma^2_a - \sum_{a \in [n]} \sum_{\ell=2}^{N_{a,h}} \frac{\sigma^2_a \big(h - f_a^{\pi}(\ell)\big)}{\ell(\ell-1)}.
\end{equation*}

\emph{Step 4: the double sum is the value.} At each time step $t \in [h]$, arm $a_t$ receives its $\ell$-th observation with $\ell = N_{a_t,t-1} + 1 \geq 2$, so $t = f_{a_t}^{\pi}(\ell)$ and $R_t(a_t) = \frac{\sigma^2_{a_t}(h - t)}{\ell(\ell-1)}$. Conversely, every observation $\ell \geq 2$ of every arm $a$ happens at exactly one time step $f_a^{\pi}(\ell) \in [h]$. Re-indexing the sum in $V^{\pi}_h$ by arm and observation count gives Equation~\eqref{eq:value}, and substituting it above proves Equation~\eqref{eq:loss-value}.
\end{proof}

Equation~\eqref{eq:value} says that the $\ell$-th observation of arm $a$, made at time $t$, contributes $\frac{\sigma^2_a (h-t)}{\ell(\ell-1)}$ to the value: observations are worth more when they are made earlier and when the arm has been observed fewer times.

\section{Optimality of the Greedy Policy} \label{sec:greedypolicy}
The main motivation for the previous section was to convert loss minimization into reward maximization so that existing methods, such as those for rotting bandits \cite{levine2017rotting}, decaying bandits \cite{heidari2016tight}, or learning in Markov decision processes \cite{azar2017minimax}, could be applied. For example, the reward of the $\ell$-th pull of arm $a$, $\frac{\sigma^2_a(h - f_{a}^{\pi}(\ell))}{\ell(\ell-1)}$, decreases as $\ell$ increases, since both $h - f_{a}^{\pi}(\ell)$ and $\frac{1}{\ell(\ell-1)}$ decrease. However, because $R_t$ depends on both the time step $t$ and the count $N_{a,t-1}$, existing results do not apply directly.

The most relevant existing result is the optimality of the greedy policy for decaying bandits \cite{heidari2016tight}, where an arm's reward depends only on how many times it has been pulled. That result does not cover rewards that depend on both the time step and the number of previous pulls of the arm, and in general the greedy policy is not optimal for such rewards. As a counterexample, consider two arms whose reward $\rho_a(\ell, t)$ depends on the pull count $\ell$ and the time step $t$, with horizon $2$ and
\begin{align*}
    \rho_1(1,1) &= 6, & \rho_1(2,2) &= 6, & \rho_1(1,2) &= 0, \\
    \rho_2(1,1) &= 7, & \rho_2(2,2) &= 0, & \rho_2(1,2) &= 0.
\end{align*}
Arm~1 yields reward at time~2 only if it was also pulled at time~1, while arm~2 yields reward only at time~1. The greedy policy selects arm~2 at time~1 and collects $7 + 0 = 7$ with either action at time~2, whereas the optimal policy $\langle 1, 1 \rangle$ collects $6 + 6 = 12$.

One could also cast the problem as a Markov decision process with the reward above. Since the reward depends on the remaining number of time steps, the state of such an MDP must include the time step, and existing regret bounds \cite{azar2017minimax} would then scale with the horizon through the size of the state space.

These challenges call for a direct analysis of the reward derived above. We now show that, for this reward, the greedy policy is optimal. Since all arms share the factor $h - t$ at time $t$, for $t < h$ the greedy action maximizes $\frac{\sigma^2_a}{(N_{a,t-1}+1)N_{a,t-1}}$, and at $t = h$ every arm has reward zero; in particular, acting greedily with respect to $\frac{\sigma^2_a}{(N_{a,t-1}+1)N_{a,t-1}}$ at every step yields a greedy policy for every horizon. When several arms attain the maximum, ties may be broken arbitrarily.

\begin{theorem} \label{th:greedy}
The greedy policy $\pi^g = \langle g_1, g_2, \dots, g_h \rangle$, where $g_t \in \arg\max_{a \in [n]} R_t(a)$ with counts computed along $\pi^g$, maximizes $V^{\pi}_h$, and hence minimizes $L(\bp, h, \pi)$.
\end{theorem}

The proof shows that any optimal policy can be transformed, one step at a time, into the greedy policy without decreasing its value. The key step is the following exchange lemma.

\begin{lemma} \label{lem:exchange}
Let $\pi = \langle g_1, \dots, g_m, i, a_{m+2}, \dots, a_h \rangle$ be a policy whose first $m$ actions agree with $\pi^g$, where $0 \leq m < h$, and let $j = g_{m+1} \neq i$. Define $\pi'$ as follows:
\begin{itemize}
    \item if $\pi$ never selects $j$ after time $m+1$, then $\pi'$ is $\pi$ with the action at time $m+1$ replaced by $j$;
    \item otherwise, let $x \in \{m+2, \dots, h\}$ be the first time after $m+1$ at which $\pi$ selects $j$, and let $\pi'$ be $\pi$ with the actions at times $m+1$ and $x$ swapped.
\end{itemize}
Then $V^{\pi'}_h \geq V^{\pi}_h$, and the first $m+1$ actions of $\pi'$ agree with $\pi^g$.
\end{lemma}

\begin{proof}
The second claim holds by construction. For the first, write $N_a = N_{a,m}$ for the counts after the first $m$ steps, which are the same under $\pi$, $\pi'$ and $\pi^g$. Since $j$ is greedy at time $m+1$, $R_{m+1}(j) \geq R_{m+1}(i)$, that is,
\begin{equation} \label{eq:greedy-choice}
    (h-m-1) \left( \frac{\sigma^2_j}{(N_j+1)N_j} - \frac{\sigma^2_i}{(N_i+1)N_i} \right) \geq 0.
\end{equation}
By Equation~\eqref{eq:value}, the value is a sum of contributions of individual observations. The policies $\pi$ and $\pi'$ select every arm other than $i$ and $j$ at the same times, so those arms contribute equally to both values, and we compare only the contributions of arms $i$ and $j$.

\emph{Case 1: $\pi$ never selects $j$ after time $m+1$.} Under $\pi'$, arm $j$ gains its $(N_j+1)$-th observation at time $m+1$, which contributes $\frac{\sigma^2_j (h-m-1)}{(N_j+1)N_j}$. Arm $i$ loses its $(N_i+1)$-th observation at time $m+1$, which contributed $\frac{\sigma^2_i (h-m-1)}{(N_i+1)N_i}$. Every later observation of arm $i$, made at some time $y$ as its $\ell$-th observation under $\pi$ with $\ell \geq N_i + 2 \geq 3$, becomes its $(\ell-1)$-th observation under $\pi'$, which changes its contribution by
\begin{equation*}
    \sigma^2_i (h - y) \left( \frac{1}{(\ell-1)(\ell-2)} - \frac{1}{\ell(\ell-1)} \right) \geq 0.
\end{equation*}
Therefore, by Equation~\eqref{eq:greedy-choice},
\begin{equation*}
    V^{\pi'}_h - V^{\pi}_h \geq (h-m-1) \left( \frac{\sigma^2_j}{(N_j+1)N_j} - \frac{\sigma^2_i}{(N_i+1)N_i} \right) \geq 0.
\end{equation*}

\emph{Case 2: $\pi$ selects $j$ first at time $x > m+1$.} In this case $m+1 < x \leq h$, so $h - m - 1 > 0$, and dividing Equation~\eqref{eq:greedy-choice} by $h-m-1$ gives
\begin{equation} \label{eq:greedy-choice-2}
    \frac{\sigma^2_j}{(N_j+1)N_j} \geq \frac{\sigma^2_i}{(N_i+1)N_i}.
\end{equation}
After time $x$, both policies have the same counts and select the same actions, so all observations after time $x$ contribute equally to both values. Since $\pi$ does not select $j$ during $\{m+1, \dots, x-1\}$, the $(N_j+1)$-th observation of arm $j$ moves from time $x$ under $\pi$ to time $m+1$ under $\pi'$, which increases the value by
\begin{equation*}
    \frac{\sigma^2_j (x-m-1)}{(N_j+1)N_j}.
\end{equation*}
For arm $i$, let $x_0 = m+1 < x_1 < \dots < x_{k-1}$ be the times in $\{m+1, \dots, x-1\}$ at which $\pi$ selects $i$, and let $x_k = x$. Under $\pi'$, arm $i$ is selected at times $x_1, \dots, x_k$ instead. So, for each $\ell \in [k]$, the $(N_i+\ell)$-th observation of arm $i$ moves from time $x_{\ell-1}$ to time $x_\ell$, which decreases the value by a total of
\begin{equation*}
    \sum_{\ell=1}^{k} \frac{\sigma^2_i (x_\ell - x_{\ell-1})}{(N_i+\ell)(N_i+\ell-1)} \leq \frac{\sigma^2_i}{(N_i+1)N_i} \sum_{\ell=1}^{k} (x_\ell - x_{\ell-1}) = \frac{\sigma^2_i (x-m-1)}{(N_i+1)N_i},
\end{equation*}
since $(N_i+\ell)(N_i+\ell-1) \geq (N_i+1)N_i$ and the sum telescopes. Therefore, by Equation~\eqref{eq:greedy-choice-2},
\begin{equation*}
    V^{\pi'}_h - V^{\pi}_h \geq (x-m-1) \left( \frac{\sigma^2_j}{(N_j+1)N_j} - \frac{\sigma^2_i}{(N_i+1)N_i} \right) \geq 0. \qedhere
\end{equation*}
\end{proof}

\begin{proof}[Proof of Theorem~\ref{th:greedy}]
Let $\pi^*$ be an optimal policy; one exists because $\Pi_h$ is finite. We show by induction on $m \in \{0, 1, \dots, h\}$ that there is an optimal policy $\pi^{(m)}$ whose first $m$ actions agree with $\pi^g$. For $m = 0$, take $\pi^{(0)} = \pi^*$. Given $\pi^{(m)}$ with $m < h$, if its action at time $m+1$ is $g_{m+1}$, set $\pi^{(m+1)} = \pi^{(m)}$. Otherwise, Lemma~\ref{lem:exchange} applied to $\pi^{(m)}$ gives a policy $\pi^{(m+1)}$ whose first $m+1$ actions agree with $\pi^g$ and whose value is at least that of $\pi^{(m)}$, so $\pi^{(m+1)}$ is also optimal. For $m = h$, we obtain $\pi^{(h)} = \pi^g$, so the greedy policy is optimal. By Theorem~\ref{th:equivalence}, it also minimizes $L(\bp, h, \pi)$.
\end{proof}

\begin{remark} \label{rem:classical}
Theorem~\ref{th:greedy} can also be seen through classical resource allocation. The term
\begin{equation*}
    \frac{\sigma^2_a}{(N_{a,t-1}+1)N_{a,t-1}} = \frac{\sigma^2_a}{N_{a,t-1}} - \frac{\sigma^2_a}{N_{a,t-1}+1}
\end{equation*}
is the decrease in the one-step loss $\sum_{a} \sigma^2_a / N_{a,t-1}$ from one more observation of arm $a$. For a fixed total number of observations, adding observations one at a time to the arm with the largest such decrease minimizes this separable convex loss \cite{fox1966discrete}; in survey sampling this is the exact integer form of Neyman allocation \cite{neyman1934two, wright2012equivalence}. Since the allocations produced in this way are nested, the greedy policy minimizes the one-step loss at every time step simultaneously, and therefore also minimizes the cumulative loss $L(\bp, h, \pi)$. The exchange argument above gives a self-contained proof of this fact in the reward formulation of Theorem~\ref{th:equivalence}.
\end{remark}

\begin{remark} \label{rem:adaptive}
The greedy action at each step depends only on $\bp$ and on the counts, which are determined by the earlier actions. The greedy policy is therefore a fixed action sequence, and Theorem~\ref{th:greedy} shows that it is optimal among all fixed action sequences, which is the natural benchmark for an agent that knows $\bp$. When the actions depend on the observed outcomes, the observations of an arm are no longer a fixed number of independent samples at each time step, so Equation~\eqref{eq:mse} need not hold for the resulting estimates; we leave the analysis of adaptive policies against this benchmark to future work.
\end{remark}

The most important consequence of Theorem~\ref{th:greedy} is that the factor $h - t$ can be dropped from the reward: all arms share it at each time step, so an agent can act greedily with respect to $\frac{\sigma^2_a}{(N_{a,t-1}+1)N_{a,t-1}}$ without knowing the horizon. It also allows the use of algorithms whose analysis relies on the optimality of the greedy policy, such as \cite{seznec2019rotting}. Next, we use this result to construct a UCB policy whose choices at each step are close to those of the greedy policy.

\section{Confidence Intervals and the UCB Policy} \label{sec:ucb}
By Theorem~\ref{th:greedy}, it suffices to act greedily with respect to the reward without the factor $h - t$. Given the history $\mathcal{H}_t$, the reward for selecting arm $a$ at time $t+1$ is
\begin{equation*}
    r_t(a) = \frac{\sigma^2_a}{(N_{a,t}+1)N_{a,t}}.
\end{equation*}
The counts $N_{a,t}$ are known to the agent, so a confidence interval on $r_t(a)$ follows from a confidence interval on $p_a$, which we obtain from Hoeffding's inequality \cite{hoeffding1994probability}: if $Z_1, Z_2, \dots, Z_s$ are independent random variables with $0 \leq Z_i \leq 1$ and $S = \frac{1}{s}\sum_{i=1}^{s} Z_i$, then for all $\eta \geq 0$,
\begin{equation*}
    \Pr\big(|S - \E[S]| \geq \eta\big) \leq 2 e^{- 2s\eta^2}.
\end{equation*}

\begin{lemma} \label{lem:ci}
Let $\hat{p}$ be the mean of $s$ independent observations of coin $a$, let $\delta \in (0,1)$, and let
\begin{equation*}
    \epsilon = \sqrt{\frac{\log(2/\delta)}{2s}}.
\end{equation*}
Then, with probability at least $1-\delta$,
\begin{equation*}
    \big|\hat{p}(1-\hat{p}) - \sigma^2_a\big| \leq \epsilon,
    \quad \text{and hence} \quad
    \frac{\sigma^2_a}{(s+1)s} \leq \frac{\hat{p}(1-\hat{p}) + 2\epsilon}{(s+1)s}.
\end{equation*}
\end{lemma}
\begin{proof}
By Hoeffding's inequality, $|\hat{p} - p_a| \leq \epsilon$ with probability at least $1 - \delta$. On this event,
\begin{equation*}
    \big|\hat{p}(1-\hat{p}) - p_a(1-p_a)\big| = |\hat{p} - p_a| \, |1 - p_a - \hat{p}| \leq \epsilon,
\end{equation*}
since $p_a, \hat{p} \in [0,1]$ implies $|1 - p_a - \hat{p}| \leq 1$. The second claim follows since $\epsilon \leq 2\epsilon$.
\end{proof}

Lemma~\ref{lem:ci} holds for a fixed number of observations. Since the counts $N_{a,t}$ are random and the bound must hold at every time step, the UCB policy uses a confidence width that grows logarithmically with $t$:
\begin{equation} \label{rhat}
    \hat{r}_t(a) = \frac{\hat{p}_{a,t}(1 - \hat{p}_{a,t}) + 2\epsilon_{a,t}}{(N_{a,t}+1)N_{a,t}}, \qquad \text{where} \quad \epsilon_{a,t} = \sqrt{\frac{\log(2t^2/\delta)}{2N_{a,t}}}.
\end{equation}
The resulting policy is summarized in Algorithm~\ref{alg:ucb}.

\begin{algorithm}[t]
\caption{UCB policy for online estimation of Bernoulli processes}
\label{alg:ucb}
\begin{algorithmic}[1]
\Require number of arms $n$, horizon $h$, confidence parameter $\delta$
\State Observe each arm once; set $N_{a} \gets 1$ and $\hat{p}_{a}$ to the observed outcome for all $a \in [n]$
\For{$t = 1, \dots, h$}
    \State Predict $\langle \hat{p}_1, \dots, \hat{p}_n \rangle$
    \State Select $a_t \in \arg\max_{a \in [n]} \hat{r}_{t-1}(a)$, computed from Equation~\eqref{rhat} with the current $N_a$ and $\hat{p}_a$ (using $t-1 = 1$ when $t = 1$)
    \State Observe $z_{a_t,t}$
    \State $\hat{p}_{a_t} \gets (N_{a_t}\hat{p}_{a_t} + z_{a_t,t}) / (N_{a_t} + 1)$; \quad $N_{a_t} \gets N_{a_t} + 1$
\EndFor
\end{algorithmic}
\end{algorithm}

\begin{theorem} \label{th:ucb}
Let $\delta \in (0,1)$, let $a_{t+1} \in \arg\max_{a \in [n]} \hat{r}_t(a)$ be the action of the UCB policy given $\mathcal{H}_t$, and let $g_{t+1} \in \arg\max_{a \in [n]} r_t(a)$ be the greedy action given the same history. Then, with probability at least $1 - 3n\delta$,
\begin{equation*}
    \sum_{t \in [h-1]} \big( r_t(g_{t+1}) - r_t(a_{t+1}) \big) \leq \sum_{t \in [h-1]} \frac{3\,\epsilon_{a_{t+1},t}}{(N_{a_{t+1},t}+1)\, N_{a_{t+1},t}}.
\end{equation*}
\end{theorem}
\begin{proof}
For each arm $a$, the outcome $z_{a,t}$ is independent of $\mathcal{H}_{t-1}$, and the decision to select $a$ at time $t$ depends only on $\mathcal{H}_{t-1}$. Hence the successive observed outcomes of arm $a$ are independent samples of coin $a$, even though the times at which they are made are random. Equivalently, we may draw, for each arm, an infinite sequence of independent samples in advance and let the $s$-th observation of arm $a$ be the $s$-th sample of its sequence; this does not change the distribution of the history. For each arm $a$ and each $s \geq 1$, apply Hoeffding's inequality to the mean of the first $s$ samples of this sequence with confidence parameter $\delta$ for $s = 1$ and $\delta/(s-1)^2$ for $s \geq 2$. By the union bound, all of these bounds hold simultaneously, for all $s$ and all arms, with probability at least
\begin{equation*}
    1 - n\left(\delta + \sum_{s \geq 2} \frac{\delta}{(s-1)^2}\right) = 1 - n\delta\left(1 + \frac{\pi^2}{6}\right) \geq 1 - 3n\delta.
\end{equation*}
On this event, for every $t \in [h-1]$ and arm $a$, take $s = N_{a,t}$. Since $s - 1 \leq t$ and $t \geq 1$, the Hoeffding radius for $s$ samples, $\sqrt{\log(2/\delta)/(2s)}$ for $s = 1$ and $\sqrt{\log(2(s-1)^2/\delta)/(2s)}$ for $s \geq 2$, is at most $\epsilon_{a,t}$. Hence $|\hat{p}_{a,t} - p_a| \leq \epsilon_{a,t}$, and so, as in the proof of Lemma~\ref{lem:ci}, $|\hat{p}_{a,t}(1-\hat{p}_{a,t}) - \sigma^2_a| \leq \epsilon_{a,t}$. Hence
\begin{equation*}
    r_t(a) \leq \hat{r}_t(a) \leq r_t(a) + \frac{3\epsilon_{a,t}}{(N_{a,t}+1)N_{a,t}}.
\end{equation*}
Since $a_{t+1}$ maximizes $\hat{r}_t$,
\begin{equation*}
    r_t(g_{t+1}) \leq \hat{r}_t(g_{t+1}) \leq \hat{r}_t(a_{t+1}) \leq r_t(a_{t+1}) + \frac{3\epsilon_{a_{t+1},t}}{(N_{a_{t+1},t}+1)N_{a_{t+1},t}}.
\end{equation*}
Summing over $t \in [h-1]$ gives the result.
\end{proof}

If the counts of all arms grow linearly in $t$, as they do under round robin, each term on the right-hand side decays at the rate $O(\sqrt{\log t}\, / t^{2.5})$. We do not prove that the counts under the UCB policy grow in this way. Note also that Theorem~\ref{th:ucb} compares the UCB action with the greedy action on the UCB policy's own history; it does not directly bound the difference between the values of the two policies.

\begin{remark}
The proof of Lemma~\ref{lem:ci} shows that the bonus $2\epsilon_{a,t}$ in Equation~\eqref{rhat} could be replaced by $\epsilon_{a,t}$, which would tighten the constant in Theorem~\ref{th:ucb} from $3$ to $2$. We keep the bonus $2\epsilon_{a,t}$, which is the one used in our experiments.
\end{remark}

\subsection{Extensions} \label{sec:extensions}
The preceding sections focus on a simple setting: $\bp$ is static, the agent selects one of $n$ arms, and a single agent makes all decisions. The main aim of this simplification was to obtain Theorem~\ref{th:greedy}, on which principled algorithms for more complex settings can be built. A formal treatment of the following extensions is beyond the scope of this paper; we outline how each could build on Theorem~\ref{th:greedy}.

\paragraph{Dynamic $\bp$.} A natural extension is when $\bp$ changes over time, which makes the reward non-stationary, as in non-stationary bandits \cite{nguyen2025nonstationary}. If $\bp$ can change without restriction, learning may be impossible. The interesting case is when the change per time step is bounded, for example by $\lambda$ under a Lipschitz assumption \cite{nguyen2025nonstationary}. A common approach in this setting is to inflate the upper confidence bound $\hat{r}_t(a)$ by a term that accounts for this drift and act greedily with respect to it.

\paragraph{Subset selection.} Another natural extension is when the agent selects a subset of $K$ arms at each time step. The main challenge is that finding the best subset by enumeration requires evaluating all $\binom{n}{K}$ subsets. When the reward is submodular \cite{krause2014submodular}, the subset can instead be formed by greedy subset selection \cite{satsangi2018exploiting}, which attains a $(1 - 1/e)$ fraction of the optimal utility and has been combined with bandit learning \cite{radlinski2008learning}.

\paragraph{Decentralized setting.} Our setting implicitly assumes a single agent that keeps track of all previous observations and confidence bounds. In surveillance or assistance tasks, however, many agents may work together in a decentralized way \cite{lauri2020multi}, and the main challenge is to coordinate them. If every pair of agents could communicate at every time step, the algorithm above could be implemented directly, but this requires a number of messages per time step that is quadratic in the number of agents. Ideally, the communication cost would be linear or even sublinear in the number of agents. A simple approach is for each agent to exchange its estimates and confidence bounds with a randomly selected neighbor at each time step; we leave more sophisticated coordination strategies to future work.

\paragraph{Other losses.} One could also consider losses other than the squared error, such as the entropy of the hidden variables.

\section{Experiments}
Our experiments examine whether greedy-style learning policies outperform simple baselines in this setting. We compare four policies:
\begin{itemize}
    \item \emph{Random}: selects an arm uniformly at random at each time step.
    \item \emph{Round robin}: selects the arms in a fixed cyclic order.
    \item \emph{$\epsilon$-greedy}: with probability $1-\epsilon$, selects the arm that maximizes the plug-in reward
    \begin{equation*}
        \bar{r}_t(a) = \frac{\hat{p}_{a,t} (1- \hat{p}_{a,t})}{(N_{a,t}+1)N_{a,t}},
    \end{equation*}
    and otherwise selects an arm uniformly at random.
    \item \emph{UCB}: Algorithm~\ref{alg:ucb}, which is greedy with respect to $\hat{r}_t(a)$ in Equation~\eqref{rhat}.
\end{itemize}
Round robin is a strong baseline for this problem because of the diminishing returns of the reward: by spreading observations evenly, it keeps every arm's count, and hence its estimation error, under control. Random selection, by contrast, occasionally leaves some arms unobserved for long periods.

\begin{figure}[t]
    \centering
    \includegraphics[width=0.49\linewidth]{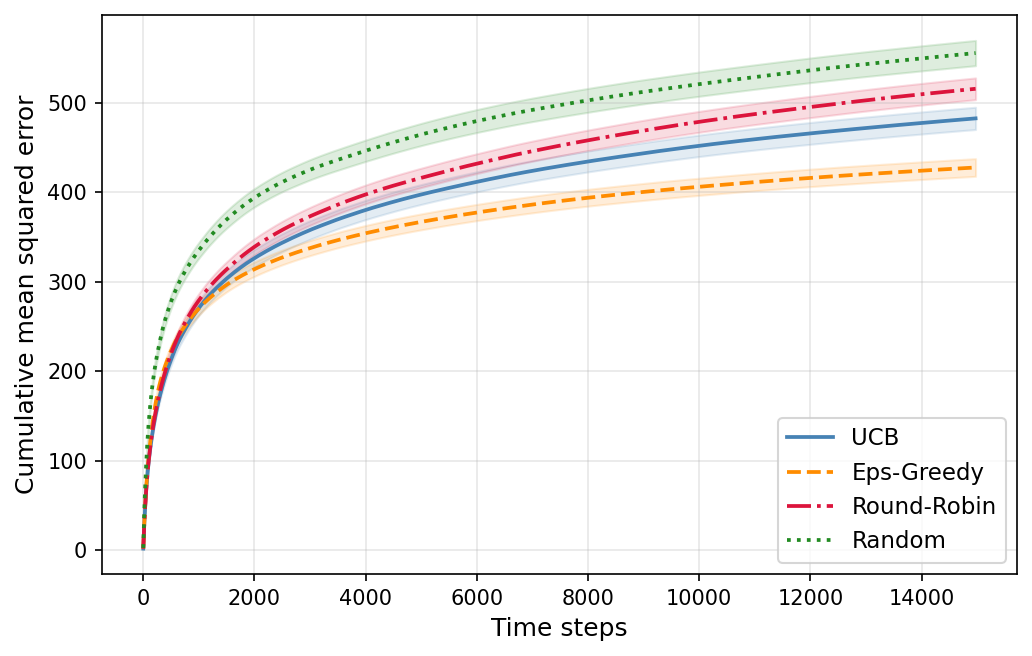}
    \hfill
    \includegraphics[width=0.49\linewidth]{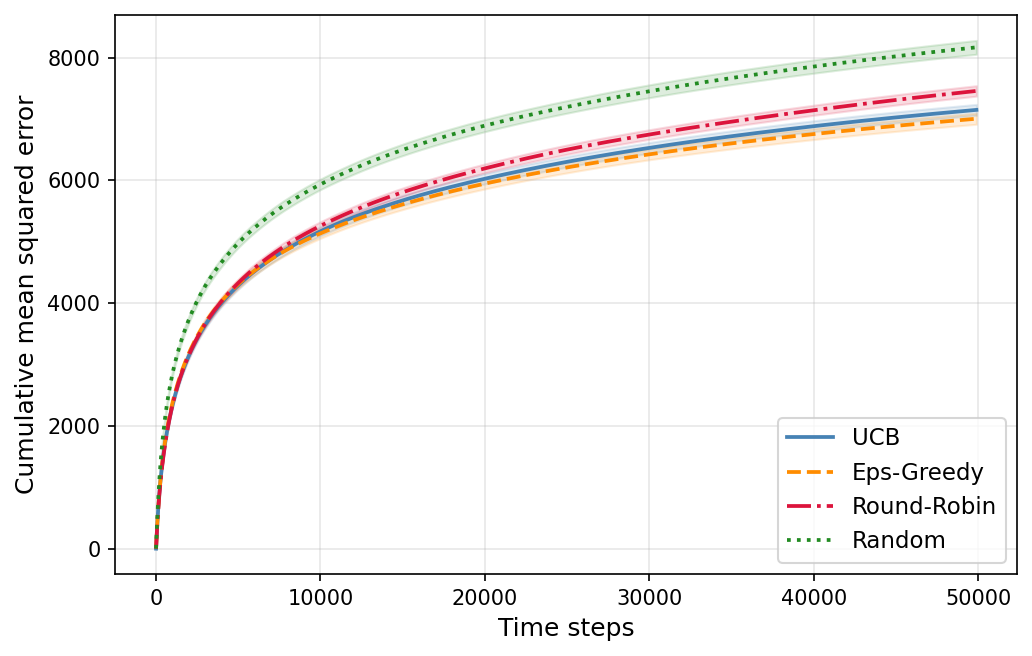}
    \caption{Average cumulative mean squared error over $100$ runs, for $\bp$ sampled at random with up to half of its entries set to $0$ or $0.5$. Left: $n = 40$. Right: $n = 100$. Shaded regions show the variability across runs.}
    \label{fig:exp1}
\end{figure}

We sample $\bp$ at random, with up to half of its entries set to $0$ or $0.5$, and repeat each experiment over $100$ independent runs. Figure~\ref{fig:exp1} shows the cumulative mean squared error between $\bp$ and its maximum-likelihood estimate for $n = 40$ arms over $15{,}000$ time steps and for $n = 100$ arms over $50{,}000$ time steps.

Both UCB (with $\delta = 0.99$) and $\epsilon$-greedy suffer less loss than the random and round-robin policies. With $\delta = 0.99$, the guarantee of Theorem~\ref{th:ucb} is vacuous, so here $\delta$ acts as a tuning parameter that controls the width of the confidence bonus. Because the confidence intervals are loose and the reward decays quickly, UCB behaves similarly to round robin early on and suffers somewhat more loss than $\epsilon$-greedy. Tighter confidence intervals and a larger number of arms may separate the two further. As expected, $\epsilon$-greedy observes arms with $p_a$ near $0$ or $1$ less often than arms with $p_a$ near $0.5$.

\section{Discussion and Future Work}
This paper presented a reward maximization formulation of the problem of estimating the biases of $n$ independent Bernoulli distributions online, and proved that the greedy policy is optimal for this reward, which allows existing ideas from online learning to be applied. Building on this result, we proposed a UCB policy and bounded, at each round, the gap between its reward and that of the greedy action. The setting studied here is simple, and, combined with loose confidence intervals and a quickly decaying reward, it leads to small differences in performance among the learning policies we studied.

In future work, we would like to extend these results to more practical settings, including non-stationary biases, subset selection, decentralized agents, and adaptive policies analyzed against the benchmark of Theorem~\ref{th:greedy}, and to develop better algorithms for the online monitoring tasks that motivate this paper.

\bibliographystyle{abbrv}
\bibliography{bernAPbib}

\end{document}